# Comparing heterogeneous entities using artificial neural networks of trainable weighted structural components and machine-learned activation functions

Artit Wangperawong[*], Kettip Kriangchaivech, Austin Lanari, Supui Lam, Panthong Wangperawong

Vody LLC, New York, NY, USA

[*] To whom correspondence should be addressed; Email: art@vody.com, artitw@gmail.com

**To compare entities of differing types and structural components, the artificial neural network paradigm was used to cross-compare structural components between heterogeneous documents. Trainable weighted structural components were input into machine-learned activation functions of the neurons. The model was used for matching news articles and videos, where the inputs and activation functions respectively consisted of term vectors and cosine similarity measures between the weighted structural components. The model was tested with different weights, achieving as high as 59.2% accuracy for matching videos to news articles. A mobile application user interface for recommending related videos for news articles was developed to demonstrate consumer value, including its potential usefulness for cross-selling products from unrelated categories.**

Documents are entities that typically contain title and content as primary structural components or fields. The title aims to be a concise summary of the entire document by highlighting its most important ideas, whereas the content describes all the ideas of the document. Documents may also contain a plethora of additional structural components, such as authors, dates, locations, summaries, credits, attributions, commentary, etc. Structural components may comprise of metadata. The structural components of a document might bear unequal weighting to achieve a given objective or to produce a certain result. Artificial neural networks can be used as a framework to calculate the optimal value of these weights [1]. To demonstrate our idea, here we describe a heterogenous document-to-document comparison system using artificial neural networks of trainable weighted structural components for the documents.

It can be useful to compare entities and documents that do not contain the same type or quantity of structural components, i.e. entities and documents that are heterogeneous. An example of heterogeneous documents are videos and news articles. How can we best determine the similarity between a document of type A and a document of type B, e.g. a video and a news article? Our methodology involves cross-comparing as many structural components of each document type as possible. For instance, a news article about a movie producer might pair well with movies by the producer as comparable documents. We intuitively expect that titles should bear more weight than contents and other components, but to quantitatively determine the magnitude of the weights we can apply the artificial neural network paradigm as shown in Fig. 1 below.

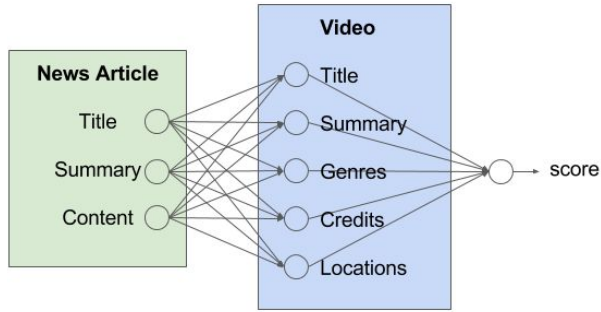

**Figure 1.** Neural network analysis paradigm for comparing heterogeneous documents, e.g. news articles and movies. Each node in the network graph is referred to as a *neuron*, which can be activated according to the structural component of the document.

Neural networks consist of weights and biases which factor into the strength of the signal or information that is propagated from one neuron (also known as nodes) to the next. To determine the value of weights and biases leading to optimal comparison, the neural network can be trained to minimize a loss function given labeled data or ground truth as inputs. The weights and biases of the network can therefore vary depending on the type of documents being compared.

The output signal of each neuron is determined by an activation function, common examples of which include the binary step-function, the sigmoid function (also known as the logistic function), the hyperbolic tangent function (tanh), and the rectified linear unit (ReLU) [2]. Activation functions are also considered to be transfer functions, and they typically need to be nonlinear to solve nontrivial problems [3]. Despite advances in artificial intelligence and deep learning involving the use of neural networks, activation functions in common practice have remained relatively simple, rote and unspecific to the application. We demonstrate in this work the non-traditional use of machine learning models as activation functions.

In the case where documents contain information in text, the vector space model of documents involving term vectors and inverted indexes is a means of document search and retrieval [4]. Search results are typically presented in descending order of a calculated match score between a single query and a single document. We apply such machine learning models as the activation functions of the neurons of the neural networks to produce scores to quantify the comparability of two documents. The scores may also be used to produce clusters of similar documents.

We designed the machine-learned activation functions to be conveniently and quickly executed on demand. The document sets of interest are preprocessed and indexed for expedient querying at a later time [5]. First, the documents are parsed into tokens using a predefined set of classes according to their structural components. Next, the tokens are converted into lexemes. A lexeme is a string that results from normalizing variations of the same word. Synonyms are added to the list of lexemes, whereas stop words--common terms that do not provide signal or distinguishable meaning--are removed [6]. Each structural component of a document can thus be represented as a term vector of such lexemes.

For better results, when possible the term frequency-inverse document frequency (tf-idf) is next calculated for each of the lexemes *t* for a structural component *d* in the document set *D* according to

$$tf \cdot idf = tf * log \frac{|D|}{1+|\{d \in D : t \in d\}|}, \textbf{(Eqn 1)}$$

where *tf* is the frequency of *t* in a given *d*, and |*D*| is the total number of documents in the set. Each structural component of a document can thus also be represented as a vector of tf-idf values for each lexeme. Finally, an inverted index is generated across all documents in each of the sets for fast querying.

In the case where we want to use $d_A$ (a single document of type A), to find the most comparable documents from $D_B$ (the set of documents of type B), the weighted term vectors $\widehat{d_{A_j}}$ of the $n$ structural components of $d_A$ can be sent into each of the activation functions of $D_B$, which calculates and outputs the weighted cosine similarity or some other metric between the two types of documents. Note that the inverted index and tf-idf values for the documents of type A would not be necessary if there is only a single document in the set. The cosine similarity measure between the a weighted structural component $d_{A_j}$ and a structural component $d_{B_j}$ is calculated as

$$\widehat{d_{A_j}} = \sum_{i}^{n} \left\{ w_{i,j} \widehat{d_{A_{0i}}} \right\} ,  \quad \textbf{(Eqn 2)}$$

$$sim(d_{A_j}, d_{B_j}) = \frac{\widehat{d_{A_j}} \cdot \widehat{d_{B_j}}}{\|\widehat{d_{A_j}}\| \|\widehat{d_{B_j}}\|} , \quad \textbf{(Eqn 3)}$$

where $w_{i,j}$ is the weight, and $\widehat{d_{A_{0i}}}$ and $\widehat{d_{B_j}}$ are the non-negative term vectors or tf-idf vectors for the $i^{th}$ structural component from the document of type A and $j^{th}$ structural component from the documents of type B, respectively. While traditional activation functions have scalar inputs, each neuron in the B (video) layer receives a vector signal. We can further restrict results to be non-negative by applying ReLUs to $sim(d_{A_j}, d_{B_j})$, $w_{i,j}$, or $\widehat{d_{A_j}}$.

The final score is calculated from a weighted summation and normalization of the resulting weighted similarity measures for each unique pair of $d_{A_j}$ and $d_{B_j}$ described in Eqn. 3 above. To additionally classify the comparison, e.g. match or no match, a simple binary step-function may be used as the activation function for the final output neuron. For the example in Fig. 1, Appendix A contains a diagram of iterating through $j$ to calculate Eqn. 2, Eqn. 3, and the final score.

In order to determine the appropriate weights for the above calculations, labeled data or ground truth can be used to calculate the error that is backpropagated throughout the network to adjust the weights according to steepest gradient descent to minimize a loss function. To classify match or no-match in the example of Fig. 1, we can use the binary cross-entropy loss [7]. For non-differentiable activation functions, the neural networks can be trained using other methods [8], such as genetic algorithms and evolution strategies [9]. Training a neural network accordingly is well established and therefore not discussed here further.

**Table 1.** News article-to-video matching accuracy for the neural network of Fig. 1 given varying weights. The weights were manually adjusted, and the accuracy was calculated from aggregating scores by human judges.

| Model | 1 | 2 | 3 | 4 | 5 |
|---|---|---|---|---|---|
| $w_{1,1}$ (title-title) | 3 | 0 | 3 | 6 | 9 |
| $w_{1,2}$ (title-summary) | 2 | 0 | 2 | 4 | 6 |
| $w_{2,1}$ (summary-title) | 0 | 3 | 3 | 3 | 3 |
| All other weights | 2 | 2 | 2 | 2 | 2 |
| Accuracy (%) | 44.2 | 41.7 | 56.7 | 59.2 | 57.5 |

For the setup shown in Fig. 1, we provide a simple example of our training results in Table 1. The weights were adjusted manually and non-exhaustively using integers. Accuracy was determined from aggregating ratings by human judges as to whether pairs of videos and news articles matched relevantly. From comparing models 1 and 2, we see that the title of the news article factors more than the news article summary in terms of accuracy. Nevertheless, the decline in accuracy in model 5 demonstrates that there cannot be an overemphasis on titles. Much higher accuracy is feasible in practice with more exhaustive training and exploration of the weight space. Additionally, the weights can be expressed as rational (floating point) numbers determined from minimizing the loss function described above. Even with such rudimentary training, useful results can be obtained, as demonstrated in the *Vody Today* mobile application user interface screenshot in Fig. 2 below.

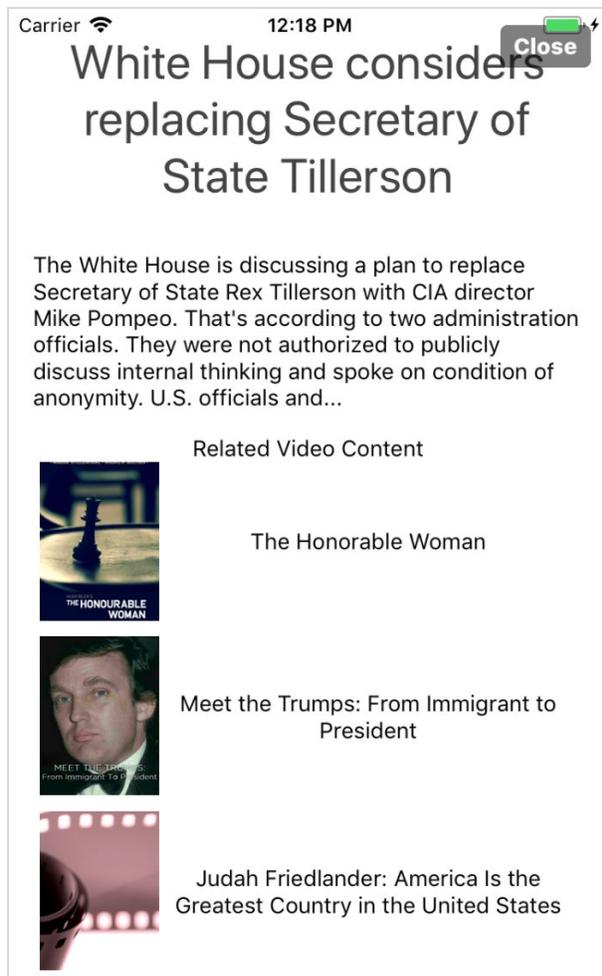

**Figure 2.** Screenshot from the *Vody Today* mobile application user interface displaying a news article (top item) and its video recommendations results (bottom 3 items). The neural network of Fig. 1 and the optimal weights (of model 4) from Table 1 are used to generate the recommendations.

Our concept of the *Vody Today* mobile app is to provide updates on current events, with movie and show recommendations based on news, location, weather, horoscopes, etc. We posit that movies and shows are most valued and enjoyed in the context of the consumer's everyday life. In terms of business, such an application can be useful for cross-selling products that are categorically different. See more screenshots of the *Vody Today* mobile app in Appendix B.

Such use of non-traditional and more customized activation functions has potential to advance the state of the art in both information retrieval and artificial intelligence. For instance, deep neural networks can be constructed to account for more than two types of documents (Fig. 3). Multidimensional neural networks can be used to cross-analyze the individual lexemes of the term vectors, which accounts for positional context. The concepts and results presented here can also be applied to non-textual data, as our use of machine-learned activation functions is compatible with common neural network computing patterns and principles, including dropout, regularization, adaptive learning, convolution, pooling, data augmentation, autoencoders, deep reinforcement learning, etc [10-12].

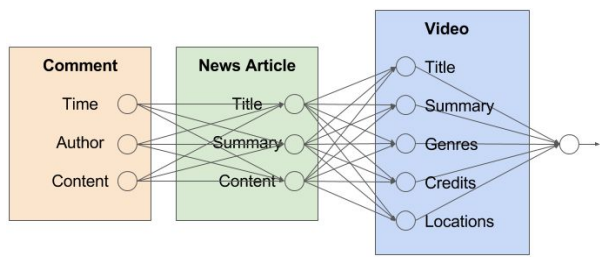

**Figure 3.** An example of a deep neural network for comparing more than two types of documents.

**Contributions**

A.W. conceived of the model, implemented the application to news articles and videos, and composed the manuscript. A.W. worked with K.K. to perform experiments to evaluate and fine-tune the model. A.W. worked with S.L. to design and build the mobile application user interface, and to procure data on news articles. K.K., P.W., and A.L. contributed to the mobile application user interface. P.W. led the effort with A.L. to procure and prepare structured data on over 1 million videos. All authors contributed ideas and reviewed the manuscript.

**Acknowledgements**

A.W. thanks Vody LLC for supporting the research and implementation.

**A. Wangperawong, K. Kriangchaivech, A. Lanari, S. Lam, P. Wangperawong.** *Comparing heterogeneous entities using artificial neural networks of trainable weighted structural components and machine-learned activation functions*, January 5, 2018.

# Appendix A

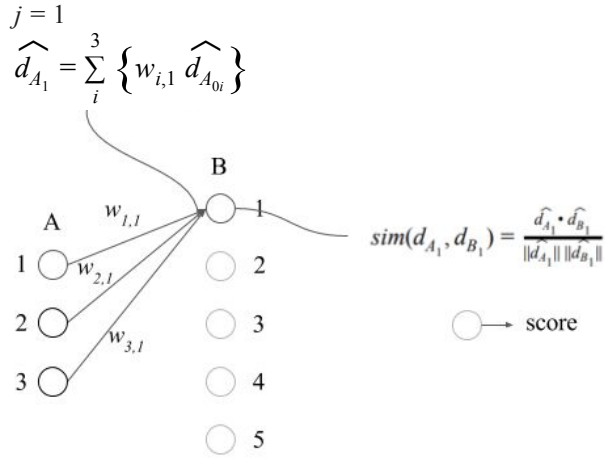

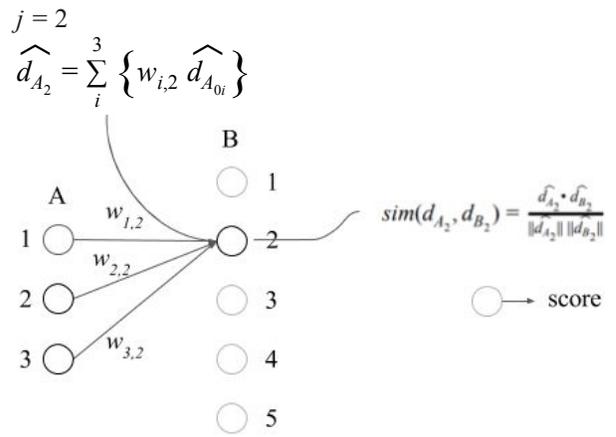

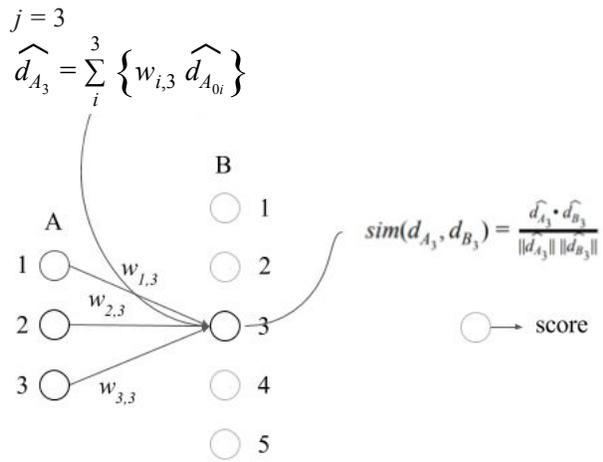

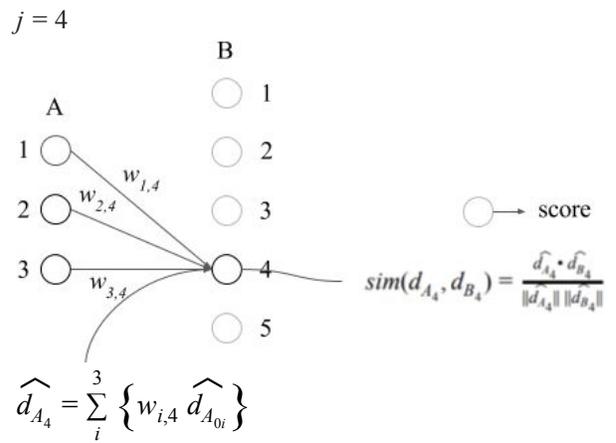

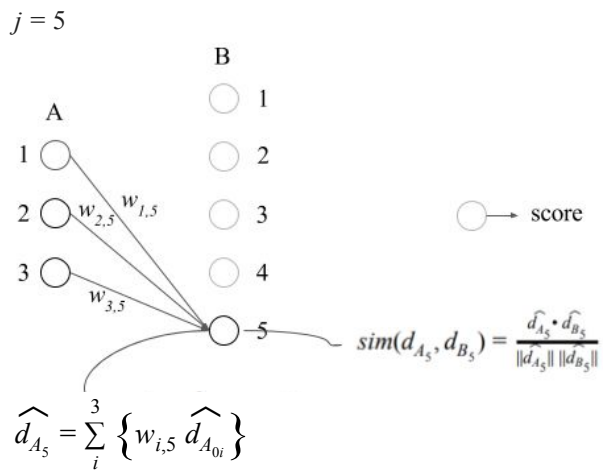

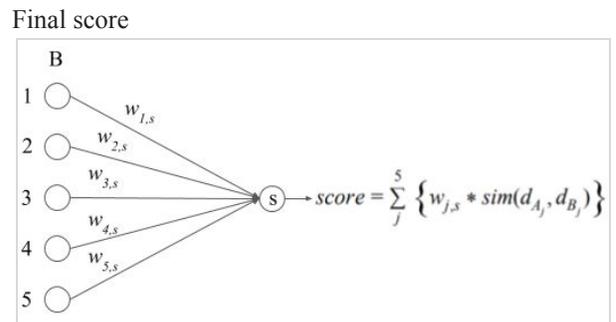

**A. Wangperawong, K. Kriangchaivech, A. Lanari, S. Lam, P. Wangperawong.** *Comparing heterogeneous entities using artificial neural networks of trainable weighted structural components and machine-learned activation functions*, January 5, 2018.

# Appendix B

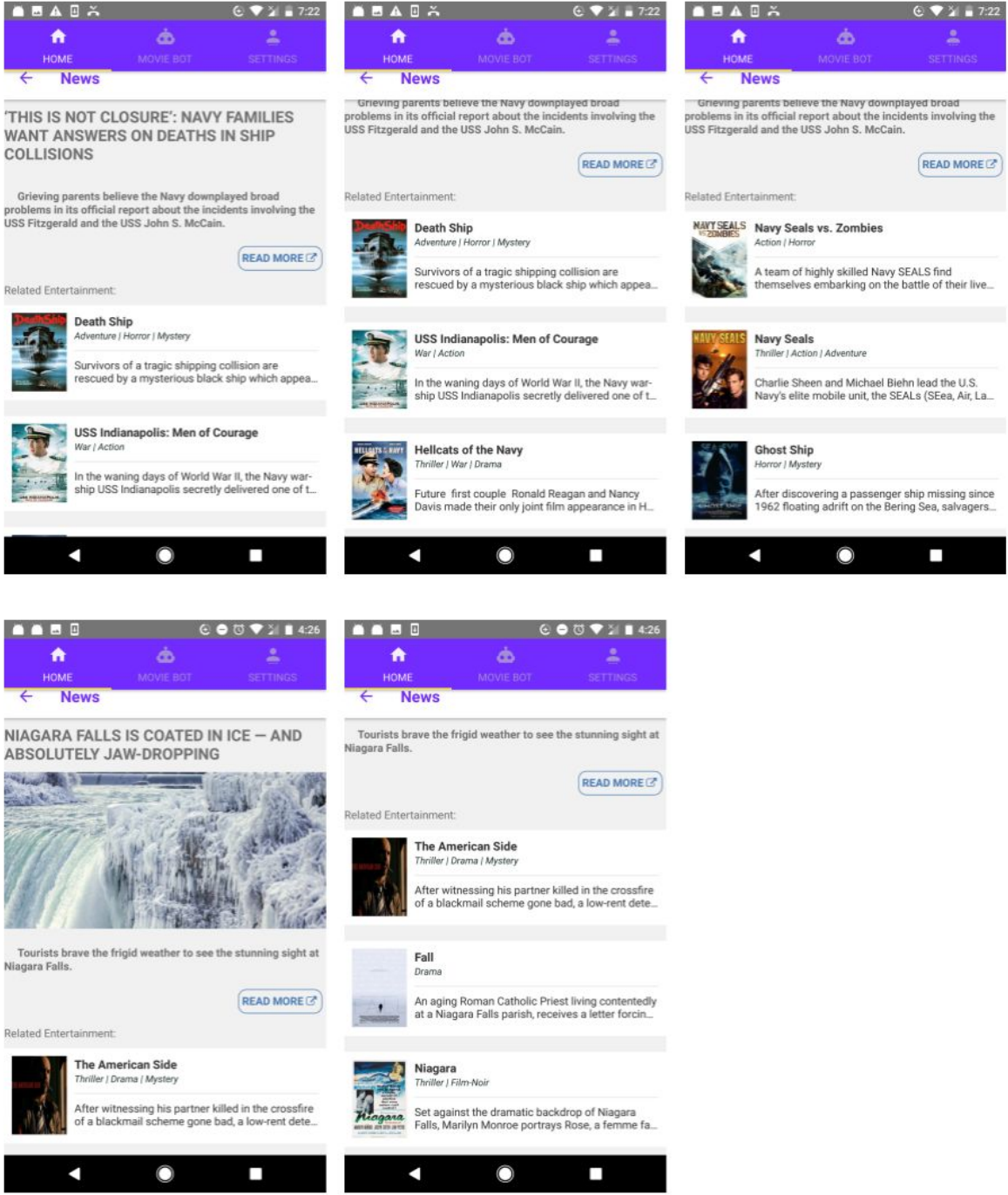

A. Wangperawong, K. Kriangchaivech, A. Lanari, S. Lam, P. Wangperawong. *Comparing heterogeneous entities using artificial neural networks of trainable weighted structural components and machine-learned activation functions*, January 5, 2018.

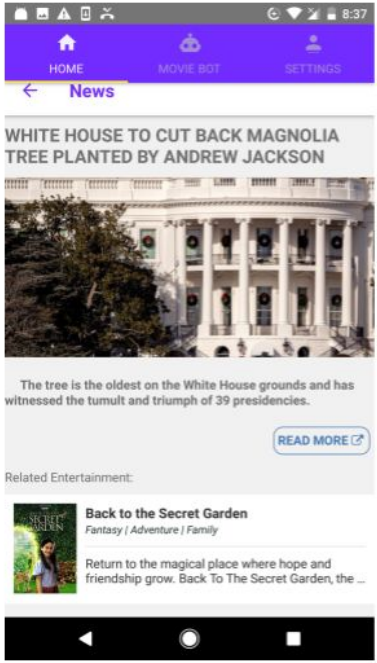
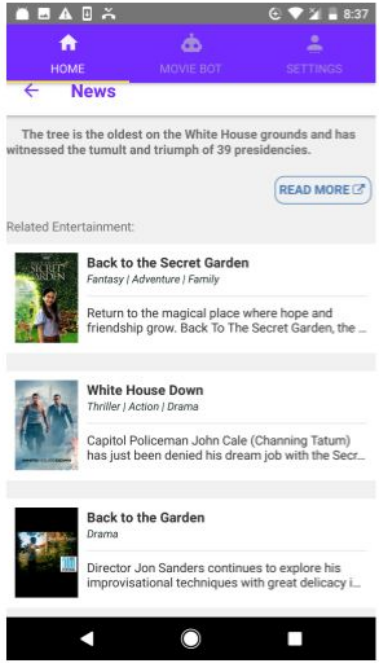
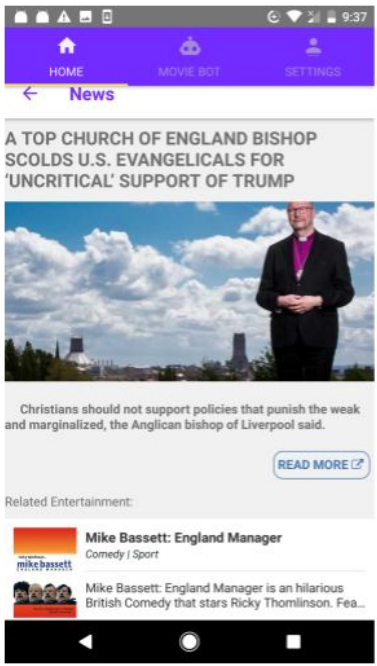
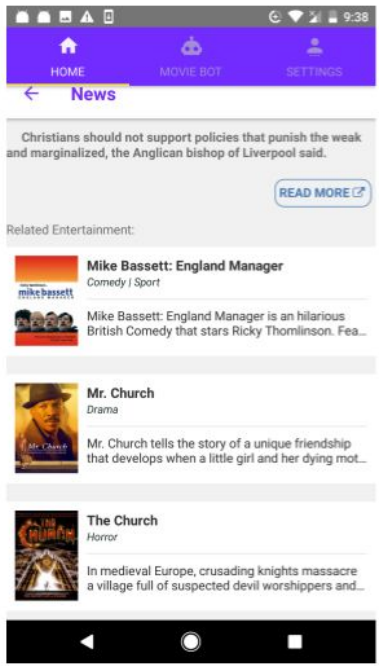

**A. Wangperawong, K. Kriangchaivech, A. Lanari, S. Lam, P. Wangperawong.** *Comparing heterogeneous entities using artificial neural networks of trainable weighted structural components and machine-learned activation functions*, January 5, 2018.

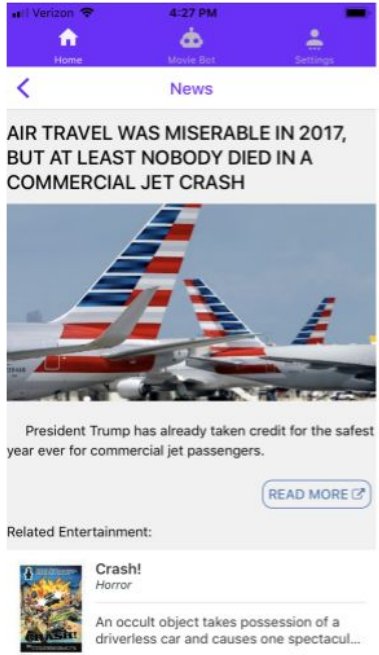
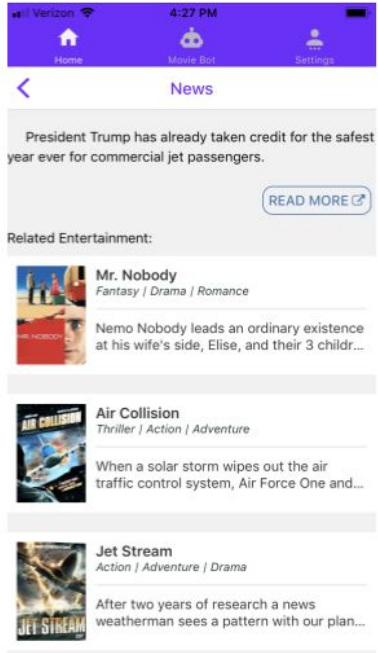
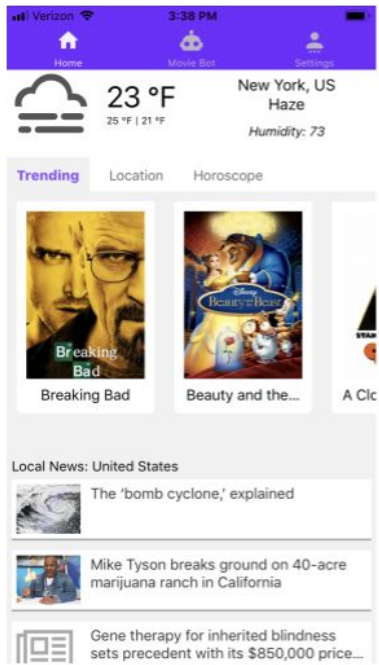
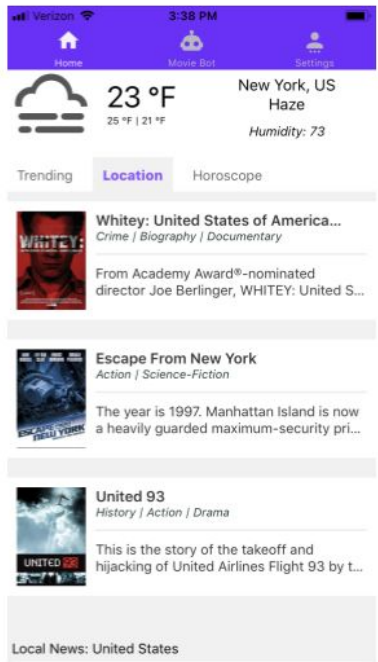
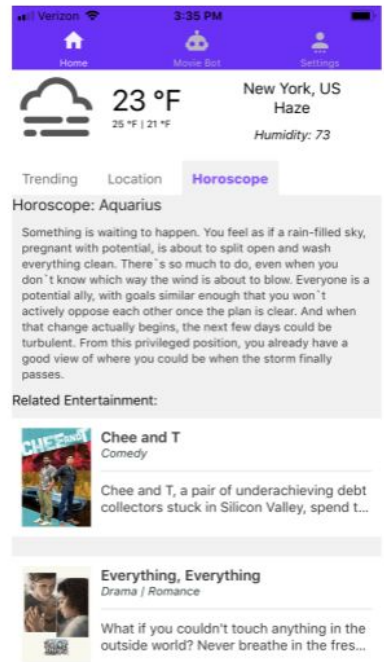

**A. Wangperawong, K. Kriangchaivech, A. Lanari, S. Lam, P. Wangperawong.** *Comparing heterogeneous entities using artificial neural networks of trainable weighted structural components and machine-learned activation functions*, January 5, 2018.